# A Model for Non-Monotonic Reasoning Using Dempster's Rule

by

Dr. Mary Deutsch-McLeish[1]
Departments of Computing Science/Mathematics
University of Guelph
Guelph, Ontario, Canada N1G 2W1
mdmcleish@water.uwaterloo.ca

Considerable attention has been given to the problem of non-monotonic reasoning in a belief function framework. Earlier work (M. Ginsberg) proposed solutions introducing meta-rules which recognized conditional independencies in a probabilistic sense. More recently an $\epsilon$-calculus formulation of default reasoning (J. Pearl) shows that the application of Dempster's rule to a non-monotonic situation produces erroneous results. This paper presents a new belief function interpretation of the problem which combines the rules in a way which is more compatible with probabilistic results and respects conditions of independence necessary for the application of Dempster's combination rule. A new general framework for combining conflicting evidence is also proposed in which the normalization factor becomes modified. This produces more intuitively acceptable results.

## 1. Introduction

In commonsense reasoning, it is often the case that one wishes to draw 'typically' information from a fact such as 'usually all $A$'s are $B$'s' or most $A$'s are $B$'s. As some exceptions to these rules do exist, they may be at best taken to be 'default' rules [11,21] or assumptions. Under the default assumption, the rules are considered to be true until this belief needs to be revised due to the arrival of contradictory pieces of evidence (such as $A$ is definitely not $B$, for some $A$.) Hence the inherently non-monotonic nature of default reasoning systems. In the original scheme proposed by Reiter, no use is made of multivalued logic, probability theory or any (quasi) numeric form of uncertainty. Many recent papers have appeared (8, 9, 10, 12, 14, 15, 16) to name but a few discussing the problem in a variety of these frameworks.

The problem of conflicting evidence and non-monotonicity in the Dempster-Shafer model of beliefs has been a topic of considerable discussion and concern in recent years. Authors such as Ginsberg [8,9], Dubois and Prade [6,7], Zadeh [25,26] and Henkind [13] have written a number of articles on the subject. Recently J. Pearl [19,20] has written articles showing presumably unresolvable difficulties with belief function theory in handling such problems.

This paper examines the issue again and offers new solutions. It begins by re-evaluating work by J. Pearl and a particular problem in non-monotonic reasoning discussed in the context of an $\epsilon$-calculus definition of default logic. When examining this in more detail, the results are shown not to be not as counter-intuitive as indicated in [19] given the assumptions made. **A new model is presented which overcomes many of the objections found in the earlier work.** Although evidence is actually combined, the results are not inconsistent with the views of Ginsberg [9]. This model is related to recent work by Dubois and Prade [6] on interpretations of dependent rules, but, in fact, is not equivalent to it. This result is then extended to provide a more general solution to problems of non-monotonicity and conflicting evidence in belief function theory. In this new framework, the often disagreeable affect of the normalizing constant in situations of conflict is overcome.

Section 1.1 provides a brief review of basic definitions. Section 2 looks in detail at the problem of evidence combination for a classical example of non-monotonic reasoning. Section 3 extends these results to more general problems of conflicting evidence and Section 4 concludes the paper.

---

[1]This work has been supported by NSERC operating grant #A4515.

## §1.1 Definitions

A brief review of a few common terms used in belief function theory are presented. [See also 4,5,22].

(1) *frame of discernment*, $\theta$, which consists of a finite set of hypotheses assumed to be mutually exclusive and exhaustive.

(2) a function $m : 2^\theta \rightarrow [0, 1]$, called a *basic probability assignment* if whenever $m(\emptyset) = 0$ and $\sum_{Y \subseteq \theta} m(Y) = 1$. The quantity $m(A)$ is a measure of belief committed exactly to $A$, where $A \subseteq \theta$, whereas the quantity $m(\theta)$ is a measure of belief that remains unassigned after commitment of belief to various subsets of $\theta$.

(3) A function $Bel : 2^\theta \rightarrow [0, 1]$; called a *belief function* if
   (i) $Bel(\emptyset) = 0$, (ii) $Bel(\theta) = 1$
   (iii) $Bel(A) = \sum_{B \subseteq A} m(B)$, for all $A \subseteq \theta$.

(4) *Plausibility* of $A = 1 - Bel(A^c)$, where c stands for complement. The interval $[Bel(A), Pl(A)]$ is called the belief interval.

(5) Suppose $m_1$ and $m_2$ are basic probability assignments over a frame of discernment $\theta$, then their *orthogonal sum* $m_1 \otimes m_2$ is defined by

$$m(\emptyset) = 0, \quad m(X) = K^{-1} \sum_{\{i,j; A_i \cap B_j = X\}} m_1(A_i) m_2(B_j)$$

where $K = 1 - \sum_{\{i,j; A_i \cap B_j = \emptyset\}} m_1(A_i) m_2(B_j)$.

Here $m = m_1 \otimes m_2$ and $X, A, B$ are subsets of $\theta$. $K$ is called the *normalizing constant*. This is Dempster's Combination rule.

*A Review of Default Reasoning using an $\epsilon$-calculus approach*

The $\epsilon$-calculus formulation in [19] presents a way in which to capture the essence of statements like "almost all $A$'s are $B$'s" by interpreting these sentences in the terms of extreme conditional probabilities, infinitesimally distant from 0 or 1. For example, the sentence almost every bird flies becomes $P(\text{Fly}(x)/\text{Bird}(x)) \geq 1 - \epsilon$, where $\epsilon > 0$ is a quantity that can be made infinitesimally small. A default theory $T =< F, \Delta >$ contains factual sentences $F$ and default statements $\Delta$. The notion of a plausible conclusion of a theory is given as follows: let $\Delta_\epsilon = \{P(q/p) \geq 1 - \epsilon : p \rightarrow q \epsilon \Delta\}$. Then $r$ is a plausible conclusion of $T$ if $P(r/F) \geq 1 - 0(\epsilon)$ where $0(\epsilon)$ represents a function of $\epsilon$ which approaches 0 as $\epsilon \rightarrow 0$. A full theory is developed tying the $\epsilon$-theory together with the work of E. Adams on the Logic of Conditionals.

Pearl examines the results of the analysis of the example of drawing a plausible conclusion from:

$$\Delta = \{\text{Penguin} \rightarrow \text{Fly}, \text{Bird} \rightarrow \text{Fly}, \text{ and Penguin} \rightarrow \text{Bird}\},$$
$$\text{and} \quad F = \{\text{Penguin(Tweety)}, \text{Bird(Tweety)}\}$$

If $\Delta_\epsilon = \{P(f/b) \geq 1 - \epsilon, P(f/p) \leq \epsilon \text{ and } P(b/p) \geq 1 - \epsilon\}$ it is shown in [19], following the usual rules of probability that $P(f/p, b) \leq 0(\epsilon)$ or equivalently that $P(\sim f/p, b) \geq 1 - 0(\epsilon)$, where $0(\epsilon) = \frac{\epsilon}{1-\epsilon}$. Thus, Tweety does not fly is a plausible conclusion of the theory. This interpretation of the classical example allows for bird $\rightarrow$ fly and penguin $\rightarrow\sim$ fly to both hold - permitting non-monotonicity.

## §2 A New Belief Function Interpretation of a Classical Example from non-monotonic Reasoning

### 2.1 Initial Comments and Comparison with Conditional Probabilities

A number of observations are made by J. Pearl in [19] concerning the use of Dempster-Shafer theory for conflicting pieces of evidence. This section provides a more complete analysis of the use of Dempster-Shafer theory for the classic example from non-monotonic reasoning concerning 'Birds normally fly'.

The discussion given in [19], begins with the following belief assignments:

$m_1 : p \rightarrow \sim f$, "Penguins don't normally fly"    rule 1
$m_2 : b \rightarrow f$, "Birds normally fly"    rule 2
$m_3 : p \rightarrow b$, "Penguins are birds"    rule 3

where $m_1 = 1 - \epsilon_1$, $m_2 = 1 - \epsilon_2$ and $1 - \epsilon_1$ and $\epsilon_2$ are small positive quantities and $m_3 = 1$. If one wishes to compute $P(f/p, b)$ using conditional probabilities, it is noted that the entailment $p \rightarrow b$ implies that $P(f/p, b) = P(f/p) = 1 - P(\sim f/p) = 1 - m_1 = \epsilon_1$. However, using D-S belief function theory the result becomes

$\text{Bel}(\text{Fly}) = \frac{\epsilon_1 - \epsilon_1 \epsilon_2}{\epsilon_1 + \epsilon_2 - \epsilon_1 \epsilon_2}$

or approximately $\frac{\epsilon_1}{\epsilon_1 + \epsilon_2}$.

The material in [19] goes on to express a number of concerns about this result. This first part of section argues that the real problem concerns the lack of independence between the two pieces of evidence used. The second part of section 4 considers the 'bird flying' example analyzed in [19], redoing the D-S analysis to combine the evidence more correctly.

The conditional probability model recognizes that $P(p \text{ and } b) = P(p)$ when writing $P(\sim f/p, b) = P(\sim f/p)$ in the 'bird' example just stated at the beginning of the section. However, the D-S formulation assumes one is combining two pieces of independent information (which is clearly not the case here). Looking at the problem of combining pieces of confirming and disconfirming evidence under a conditional independence assumption, we have $P(e_1 \text{ and } e_2/H) = P(e_1/H)P(e_2/H)$, from which $P(H/e_1 \text{ and } e_2) = \frac{P(H/e_1)P(H/e_2)}{P(H)}$ by Bayes' rule and independence of $e_1$ and $e_2$.

Assuming $P(H/e_1) = \epsilon_1$ and $P(H/e_2) = 1 - \epsilon_2$,

we have $P(H/e_1 \text{ and } e_2) = \frac{\epsilon_1(1-\epsilon_2)}{P(H)}$

and $P(\sim H/e_1 \text{ and } e_2) = \frac{(1-\epsilon_1)\epsilon_2}{P(\sim H)}$

In the D-S model, Bel(H) given the same two pieces of evidence is $\frac{\epsilon_1(1-\epsilon_2)}{K}$ and $\text{Bel}(\sim H)$ is $\frac{\epsilon_2(1-\epsilon_1)}{K}$, where $K = 1 - \text{Bel}(\phi) = \epsilon_1 + \epsilon_2 - \epsilon_1\epsilon_2$.

Unlike the probability model, $\text{Bel}(H)$ and $\text{Bel}(\sim H)$ do not need to sum to 1, although they almost do here as $\text{Bel}(\phi) = \frac{\epsilon_1 \epsilon_2}{K}$ is the only other non-zero belief.

If one examines these expressions, in the D-S setting it was considered paradoxical in [19] that $\text{Bel}(H) \rightarrow 1$ if $\epsilon_2 \rightarrow 0$, independently of $\epsilon_1$. In the example it was applied to there it is more likely that $\epsilon_1$ will be smaller than $\epsilon_2$. If $\epsilon_1 = \epsilon_2^2$, for example (there are a lot fewer flying penguins than there are non-flying birds), $\text{Bel}(H) = \frac{\epsilon_2}{1+\epsilon_2} = \epsilon_2$, a small quantity. If $\epsilon_2 \rightarrow 0$, then there are really no non-flying birds, which include penguins and it does not really seem surprising that $\text{Bel}(H) \rightarrow 1$. **The two pieces of evidence, (although treated independently) are in fact not. At least in the interpretation stage this should be remembered, if not earlier as discussed in the next section.**

If $\epsilon_1 = \epsilon_2 = \epsilon^2$, $\text{Bel}(H) = 1/2$. In this situation, the evidence for and against the hypothesis are equally balanced and the belief becomes $\frac{1-\epsilon}{2-\epsilon}$ or approximately $1/2$. This result is similar to what would be obtained from a MYCIN certainty factor model[2]. Otherwise if the epsilon's are not equal it is the relative size of $\epsilon_1$ and $\epsilon_2$ that is important. If $\epsilon_1 < \epsilon_2$ (there is stronger disconfirming evidence), $\text{Bel}(H) < 1/2$ and visa versa otherwise - all of which is quite reasonable if it is assumed that we are considering two independent pieces of evidence when doing the combination.

When considering the probability model, $P(H/e_1 \text{ and } e_2)$ can become large if the prior was small and vice versa for $P(\sim H/e_1 \text{ and } e_2)$. Thus the evidence which is very different from the prior has the stronger effect on the combined conditional probability. Under the independence assumption, $P(H/e_1 \text{ and } e_2)$ is certainly not $\epsilon_1$. It is the expressions with the same numerator as the D-S expressions but different denominators which should be compared. The paper[17] by this author discusses the conditions for the equality of these two expressions in some detail.

## 2.2 The Problem of Choosing a Frame of Discernment

This section suggests a way of looking at Dempster-Shafer theory which more closely resembles the probabilistic result and captures the fact that penguins form a sub-class of birds. In fact there is inherently a problem with the translation of the problem to the D-S model in any case, as the outcome spaces are not the same. That is the value $m_1$ has been obtained as a measure over the outcome (event) space of flying and non-flying penguins, whereas $m_2$ is over the space of all birds. One is then asked to combine evidence about events where the measure of belief has been originally assumed to be spread over these different spaces. One then really assumes (in what has been done in [19]) that the value pertaining to the set of all birds also pertains to the penguin outcome space and then the evidence is combined over that space. The measure $1-\epsilon_2$ is assigned to Bel(penguins fly) and it is not too surprising to discover that Bel(fly) tends to 1 if $\epsilon_2$ goes to zero. This is actually similar to doing a Dempster conditioning (see [24]) on the space of birds (conditioning on penguins). If the belief is viewed as coming in any way from statistical evidence, the conditioning or inheritance is quite invalid. Any statistical evidence about birds flying in general implies nothing about the frequency with which penguins fly. Even if one takes a subjective view of belief, it is still unclear that this conditioning makes sense. (More will be said about this later.)

In any case, if the conditioning is carried out, one obtains for the same evidence, two different mass functions. One can hardly agree to combine them as if coming from independent pieces of evidence - at least not in the simple interpretation of the rules as evidence $\rightarrow$ outcome.

The non-monotonic problem has been studied and some resolutions have been proposed by other authors, although these suggestions have not been considered in [19]. In particular, Ginsberg [9] introduces metarules such as "never apply a rule to a set when there is a corresponding rule which can be applied to a subset". This would imply we should find the same result as for conditional probabilities, that is, only use the rule about penguins and find the belief that they fly is $\leq \epsilon_1$ (assuming Bel(doesn't fly) is $1-\epsilon_1$ and $m(\theta)=\epsilon_1$).

However, if one wishes to have a framework in which to capture both pieces of knowledge and with which to reason about future information about, perhaps, non-penguins and penguins, a way can be found to incorporate the new evidence about penguins. Bhatnagar and Kanal [1] have noted similar problems when they say that "updating of the $m$ values in the light of new evidence poses another problem ... the new evidence may require creation or deletion of focal elements or readjustment of $m$ values for the same core of evidence." In the problem being studied, the evidence for the class of penguins may be considered as new evidence from the original evidence about birds in general. In another sense, the problem is that the frame of discernment has now changed. As noted by the same authors in [1], "another constraint that exists for Dempster's rule is that the two frames must be identical ... a body of evidence for a given world model may consist of a number of frames of discernment ... combination of such bodies of evidence is an interesting and complex problem".

## 2.3 Solution to the Frame Problem.

The simple frame of discernment chosen as 'flies', 'not flies' in [19] is a very coarse frame. The question is 'what' flies or what is the variable $q$ whose possible values $q$ represents? In one case, it seems to be penguins and for the other rule it is birds.

This paper will consider a way to produce a measure $m$ and corresponding belief function over an ap-propriate frame of discernment which allows the two rules to be expressed by one function. If another rule is added to the system which is independent from the others and can be expressed over the same frame of discernment, it may be combined with this function in the usual manner for D-S theory.

**The New Frame of Discernment:** The natural frame of discernment here is the set of all birds, classed in two groups, flying birds and non-flying birds. This is essentially that used by Pearl in [19]. However to make the rule about penguins fit into the model, this frame needs to be refined. Recall from [22] that a refinement occurs when one frame of discernment $\Omega$, is obtained from another frame $\Theta$ by splitting some or all of the elements of $\Theta$. Let $w(\{\theta\})$ represent those possibilities into which $\theta$ has been split. The sets $w(\{\theta\})$ must constitute a disjoint partition of $\Omega$. Here we will split flying and non-flying birds as follows:

$w(fb)$ = {flying penguins, other species of flying birds}
$w(nfb)$ = {non-flying penguins, other species of non-flying birds}

(Note, the 'other species' may be itemized or grouped). Then $\Omega = \{fp, nfp, ofb, onfb\}$ is a refinement of $\theta = \{fb, nfb\}$, where for example, $ofb$ stands for other flying birds.

Suppose we are given a belief function $Bel_0$ on $2^\theta$. Then this is consistent with a belief function $Bel$ on $2^\Omega$ if $Bel_0(A) = Bel(w(A))$ for all $A \in 2^\theta$. In our case, we are given a belief function on $2^\theta$ by rule 2; specifically $m(fb) = 1 - \epsilon_2$, $m(\theta) = \epsilon_2$. If we let $Bel\{fp, ofb\} = 1 - \epsilon_2$, we will have a consistent belief function on $2^\Omega$.

The problem then becomes how to interpret rule 1 in this refined frame. The following Theorem provides a solution.

## Combination of Evidence

**Theorem 1.** If a refined frame $\Omega = \{fp, nfp, ofb, onfb\}$ replaces the coarser frame (which was essentially $\{flies, \sim flies\}$), and new mass functions are appropriately specified, Dempster rule of combination of the two interpreted rules results in a new belief function $Bel'$ with the following properties:

$bel'\{fp, ofb\} = 1 - \epsilon_2$, the belief interval for $\{fp\}$ is $[0, 1 - Bel'\{fp\}^c] = [0, \epsilon_1]$.

**Proof**

Consider the following table, using the usual Dempster rule of combination.

|  | $\{fp\}^c\ 1 - \epsilon_1$ | $\Omega\ \epsilon_1$ |
|---|---|---|
| $\epsilon_2 \Omega$ | $\epsilon_2(1 - \epsilon_1) : \{fp\}^c$ | $\epsilon_1 \epsilon_2 : \Omega$ |
| $\{fp, ofb\}\ 1 - \epsilon_2$ | $(1 - \epsilon_1)(1 - \epsilon_2) : \{ofb\}$ | $\epsilon_1(1 - \epsilon_2) : \{fp, ofb\}$ |

This results in a new belief function, say $Bel'$, where $Bel'\{fp, ofb\} = \epsilon_1(1-\epsilon_2)+(1-\epsilon_1)(1-\epsilon_2) = 1-\epsilon_2$, which agrees with $Bel$ on $\Omega$. There is no mass on $\{fp\}$. However the belief interval is $[0, 1 - Bel'\{fp\}^c] = [0, \epsilon_1]$. This is consistent with the model of conditional probabilities. We see that the belief interval for $\{ofb\}$ becomes $[(1-\epsilon_1)(1-\epsilon_2), 1]$, which is reasonable. The belief in $\{fp\}^c$ becomes $1 - \epsilon_1$, as before. This is really the belief that other birds fly. We cannot expect to obtain a mass on $\{fp\}$ alone as the evidence in the rule was originally given only in terms of not flying. Strictly speaking, we are not allowed to assume this implies $m(fp) = \epsilon_1$ in the D-S model. $Bel'$ is then a new belief function which combines both rules and maintains their separate validity.

Note: Many of the ideas in the first part of this paper have been expressed initially in the paper by McLeish [18].

Another way to word the members of the frame of discernment $\Omega$ would be as:
{penguins fly, other birds fly, ...other hypotheses}.

This interpretation then makes the belief $(1-\epsilon_1)(1-\epsilon_2)$ obtained for other flying birds more meaningful as the belief that other birds (not penguins) do fly. To use the new belief function, if one is given only the evidence bird and wishes to know the belief that birds fly, one considers $Bel'\{fp, ofb\}$ and if the evidence is penguins the subsets involving penguins apply.

It is very important to recognize that several transformations have taken place.

(1) A refinement of the frame (which is necessary to house both rules regardless if one is considering beliefs as being obtained subjectively or statistically). See also Shafer [23], who mentions the possible need of a refined frame for this problem.

(2) The information has been translated into a form where we consider that we have two independent pieces of evidence: $e_1$ which says the belief that penguins don't fly is $1-\epsilon_1$ and $e_2$, which says that the belief birds fly is $1-\epsilon_1$. That is, some evidence has been obtained to substantiate the outcome "birds fly" and some "other" evidence has been used to discover the information about penguins. For example, the evidence could come from observing groups of birds and then a collection of penguins - or just from someone telling you their subjective opinion. (Even if it is coming from the same individual, their way of arriving at both the outcomes might well be different.)

If the evidence is the same, for example, if one observes a population of birds and uses the same distribution for penguins, again the evidence should not be combined.

(3) If you wish to combine further new evidence, such as penguins fly with belief .2, you simply use Dempster's rule with $m(\Omega) = .8$, $m(pf) = .2$ and the table in Theorem 1. One obtains a belief that birds fly of $\frac{.8(1-\epsilon_2)}{.8+.2\epsilon_1} \cong (1-\epsilon_2)$, but is very slightly smaller. The belief that penguins fly becomes $\frac{.2\epsilon_1}{.8+.2\epsilon_1}$ or slightly more than 0. These are the results of combining all three rules and are not unreasonable.

## Comments on the Chosen Frame and Rule Interpretations

(1) **Relationship to the Work of Dubois and Prade [7].**

In a recent paper by Dubois and Prade [6], a method is proposed to handle possibly dependent rules of the form:

If $v_1 \in B_1$, then $x \in A_1$

If $v_2 \in B_2$, then $x \in A_2$, where $x$ is a variable ranging on $\Omega$ and $B_1 \subseteq V_1$, $B_2 \subseteq V_2$.

The Cartesian product $V \times \Omega$ becomes the space over which mass assignments are made (i.e., the frame of discernment). If $\alpha$ is understood to be the strength of the rule, then the mass function assigned is $m(B \times A^c)^c = \alpha$ and $m(V \times \Omega) = 1 - \alpha$. However in order to pursue this further for the problem being studied, one would already need to assume rule 1 was given in the form, penguins fly with strength $\epsilon_1$, so that $A_1 \cap A_2 \neq \emptyset$; that is the two rules must both pertain to flying. Some results for handling the situation are given when the rules are certain. When they are uncertain and there is dependence, they state that "the use of Dempster rule for combining the rules pertaining to $A_1$ and $A_2$ is forbidden". Again (as in Ginsberg [8]) neglecting a rule is suggested as a remedy in some situations.

However, the interpretation used in Theorem 1 can be viewed at least partially in the context of the approach in [7]. We let $V_1 = V_2 =$ all birds and $\Omega$ is flies, not flies. Rule 1 has $B =$ penguins, $A =$ doesn't fly and $(B \times A^c)^c$ is then $(fp)^c$ which has mass $1-\epsilon_1$, $m(V \times \Omega) = \epsilon_1$. This interpretation of a negative statement agrees with examples in [2] and [22] (where evidence against a hypothesis is put on the complement of the hypothesis). It also obeys a principle of minimum specificity (see [7] etc.). The second rule isn't modeled by this method and indeed violates $A_1 \cap A_2 \neq \emptyset$ ($A_1 =$ doesn't fly, $A_2 =$ fly). This rule is translated directly as $m(B \times A) = 1 - \epsilon_2$ and not as $(B \times A^c)^c$. More specific information is used.

(2) **A comment on the interpretation of the rules:**

The interpretation of $p \rightarrow \sim f(1 - \epsilon_1)$ as $m((pf)^c) = 1 - \epsilon_1$ is standard for the interpretation of negative evidence in Dempster-Shafer Theory [2,22]. For the information $b \rightarrow f$, suppose one wanted to examine a strict logic interpretation; $\sim b \vee f$. For a start, these events are not mutually exclusive. It would make sense then to consider a mass on $(\sim b \vee fb)$. If this is assigned the value $1 - \epsilon_2$, what does this mean? Does it mean usually one can't tell the difference between a flying bird and a non-bird, but the outcome is likely one or the other. Is this really what the rule is saying? It's more likely saying that $Bel\{\sim b \vee fb\}$ is $1 - \epsilon_2$ and $m(fb) = 1 - \epsilon_2$. (There is really no evidence to support $\sim b$). Usually positive evidence for an outcome is treated as simply putting the mass on that outcome - not as putting some mass on the possibility that the evidence is negative. In a probabilistic interpretation, $p(f/b)$ and not $p(\sim b \vee f)$ is used to model $b \rightarrow f$ (the two will be the same if $p(\sim b)$ is assumed to be zero). If one does obtain later information about whether or not non-birds fly, then $\Omega$ could be expanded to include these hypotheses and the masses over the enlarged frame would need to be readjusted. In the framework of Theorem 1, $m\{\sim b, ofb, fp\} = 1 - \epsilon_2$; the belief in $fp$ is unchanged and the masses on other sets are the same with these sets modified to include $\sim b$.

### (3) Further Comments on Conditioning

If any combination of evidence is to be done at all for the problem, as mentioned before the frames need to be merged. One way is to downward condition (as mentioned earlier) and this is effectively what has been done by J. Pearl in [19]. The other is to refine the space of birds to include penguins. However then the rule about penguins becomes interpreted in a less specific fashion when the information must be spread over a larger frame. If you combine the evidence in the larger frame and then attempt to condition down again to the smaller one, the same results are obtained as if one first conditioned downwards from the larger frame and then combined the evidence. Thus one either accepts or rejects the concept of conditioning the information about birds on the subspace, penguins - whenever it is carried out is immaterial.

As mentioned earlier, if one takes a more frequentist (random set) interpretation of belief functions, the Dempster conditioning is not justified in this situation. However one could argue that in the interpretation used here for the rules, it is not justified either. For one thing, knowing you are considering penguins doesn't provide a rule about whether they fly. (We are assuming we have evidence, say $e$, which supports rules about whether or not birds or penguins fly.) The fact that you are considering penguins does not provide such a rule. Knowing you have a penguin, over the original space {penguin's fly, penguin's don't fly} would give you a vacuous belief function ($m(\Omega) = 1$). Over the larger frame it should also. If you say something else, like $m\{$penguin's fly, penguin's don't fly$\} = 1$, and $m(A) = 0$ for all remaining sets, then this is actually saying there is evidence that would support penguins flying and evidence to support that they don't and its never possible to distinguish non-flying from flying but they all do one or the other. This in fact contradicts the first rule. It's also saying the evidence indicates there is no possibility that other birds fly. Actually the fact that one is interested in penguins says nothing about whether other birds fly - and so it is really only the vacuous belief function which makes sense.

One might wish instead to assign a belief of 1 to $\{fp, \sim fp\}$. This still has the problem that it is saying that you have no belief that other birds fly. However, if you take this and condition on beliefs alone (not masses), the result of Theorem 1 will be that penguins don't fly will be $1 - \epsilon_1$ and that they do will remain zero.

## Discussion of Other Interpretations

If one is allowed to assume from rule 1 that $m(fp) = \epsilon_1$, we obtain a situation much like that of the cabbage seed discussed by Shafer in Example 4.2 [22]. Here one piece of evidence supports the hypothesis that a plant is a cabbage and another that it is actually a dictotyledon. Cabbages are a subset of dictotyledons. If $A = \{$cabbage$\}$, $B = \{$dictotyledons$\}$ and $\Theta$ is the set of species of plants, one obtains $Bel(A) = S_1$, $Bel(B) = 1 - (1 - S_1)(1 - S_2)$, where $S_1$ and $S_2$ are the supports on $A$ and $B$ initially and Bel represents the result of forming the orthogonal sum.

In our example, if $m_1(fp) = \epsilon_1$ and $m_2(fb) = 1 - \epsilon_2$ over the frame of all birds one obtains $Bel(fp) = \epsilon_1$, $Bel(fb) = 1 - \epsilon_2 + \epsilon_1\epsilon_2$. This maintains the value of $Bel(fp)$, but increases the value of $Bel(fb)$. The assumption about rule 1 necessary to make this interpretation may not be considered valid because it is taking a probabilistic view of negation. However, it is mentioned to indicate that there are examples of subclasses which are discussed by Shafer in [22].

Yet one more interpretation might be considered. Suppose we assign $m\{nfp\} = 1 - \epsilon_1$ and $m(\Omega) = \epsilon_1$ for rule 1. Then we obtain $Bel\{nfp\} \cong \frac{\epsilon_2}{\epsilon_1+\epsilon_2}$, which approaches 1 as $\epsilon_1 \to 0$, as expected. There is no information about the subset flying penguins alone ($Bel\{fp\} = 0$). Bel(birds fly) $\cong \frac{\epsilon_1}{\epsilon_1+\epsilon_2} \to 1$ as $\epsilon_2 \to 0$ as expected. This result could be thought of as a type of partial conditioning, where $m(fp)^c$, conditioned on penguin, becomes $m\{nfp\} = 1 - \epsilon_1$. But $m(\Omega)$ remains as $\epsilon_1$. It is interesting to note that $\frac{\epsilon_2}{\epsilon_1+\epsilon_2} \leq 1 - \epsilon_2$ and $\frac{\epsilon_1}{\epsilon_1+\epsilon_2} \leq 1 - \epsilon_1$. Thus, the strength of the rules has been modified slightly in the combination process. (Actually the modification could be significant; if $\epsilon_1 = \epsilon_2$ for example, these expressions become $1/2$). These expressions (as examined earlier in the paper) are very sensitive to the relative size of the epsilons.

Of all the above, methods presented in Theorem 1 and its discussion are consistent with an example in [2] of conflicting evidence where the opposite of a value is interpreted as the complement of the set representing it. It produces a new belief function which is really a redefinition of the two separate ones over a refined frame of discernment.

## §3. Extension to Other Problems of Conflicting Evidence

This section considers some simple situations involving conflicting rules (or a default rule contradicted by a new rule). The method for processing such a system of rules proceeds in two steps. The first step incorporates the rules under one frame of discernment. The second step recombines this resulting evidence in coarser frame. The first process is similar to what has been done in Section 2. However the rules used there had the added problem that one piece of evidence was a subclass of the other and they were not independent. Here we assume we don't have any additional information about how the pieces of evidence might be related. If we suppose that we have to rules of the following form:

$$a_1 \to b(1 - \epsilon_1)$$

$$a_2 \to \sim b(1 - \epsilon_2)$$

We might also suppose that $a_1 \to b$ is a default rule and that later a new piece of evidence $a_2$ is found which implies a contradiction to this.

**Theorem 2.** Using an interpretation of rules as described in Section 2 where the frame of discernment $\Omega$ becomes the product space $\{(a_1, b), (a_1, \sim b), (a_2, b), (a_2, \sim b)\}$, the following results are obtained after performing Dempster's combination rule:
1) The belief interval for $\{(a_1, b)\} = [1 - \epsilon_1, 1]$
2) The belief interval for $\{(a_2, b)\} = [0, \epsilon_1\epsilon_2]$

**Proof:**
The combination table is slightly different from that obtained in Theorem 1 because penguins were a subclass of birds. Now the following table results:

|  | $\{a_2, b\}^c (1 - \epsilon_2)$ | $\Omega\ \epsilon_2$ |
|---|---|---|
| $\Omega\epsilon_1$ | $\epsilon_1(1 - \epsilon_2) : \{(a_1, b)(a_1, \sim b)(a_2, \sim b)\}$ | $\Omega : \epsilon_1\epsilon_2$ |
| $\{a_1, b\} 1 - \epsilon_1$ | $(1 - \epsilon_1)(1 - \epsilon_2) : \{(a_1, b)\}$ | $(1 - \epsilon_1)\epsilon_2 : \{(a_1, b)\}$ |

From this the belief intervals can be found.

If the belief functions are taken from the table of Theorem 2 and used as mass functions for the cases $b$ and $\sim b$ separately, one obtains the following theorem:

526

**Theorem 3.** If the results from the joined (product) space are now used individually to find a truly combined belief in $b$, the results become:

(1) $Bel\{b\} = \frac{\epsilon_2(1-\epsilon_1)}{(1+\epsilon_2-\epsilon_1\epsilon_2)} = O(\epsilon)$.

(2) $Bel\{b\}^c = \frac{1-\epsilon_1\epsilon_2}{1+\epsilon_2-\epsilon_1\epsilon_2} = 1 - O(\epsilon)$,

where $\epsilon$ stands for any function of $\epsilon_1, \epsilon_2$ which goes to zero as $\epsilon_1, \epsilon_2 \to 0$.

**Proof:** Now the mass functions to be combined are: $Bel'\{(a_2,b)\}^c = m_1\{b\}^c = 1 - \epsilon_1\epsilon_2$, $m_1(\theta) = \epsilon_1\epsilon_2$, and $Bel'\{(a_1,b)\} = m_2\{b\} = 1 - \epsilon_1$, $m_2(\theta) = \epsilon_1$. The resulting $m_1 \otimes m_2 = m'$ becomes:
$m'\{b\}^c = (1-\epsilon_1\epsilon_2)\epsilon_1$, $m'(\theta) = \epsilon_1^2\epsilon_2$,
$m'\{\emptyset\} = (1-\epsilon_1\epsilon_2)(1-\epsilon_1)$, $m'\{b\} = (1-\epsilon_1)(\epsilon_1\epsilon_2)$.
After normalizing one obtains the stated results.

**Corollary 3:**
If the prior probability of $b$ is non-infinitesimal, then the above result for $Bel\{b\}$ from Theorem 3 is consistent with the probabilistic result.

**Proof:** $P(b|a_1 \text{ and } a_2) = \frac{\epsilon_2(1-\epsilon_1)}{P(b)} = O(\epsilon)$ if $P(b) \gg \epsilon_2$. Indeed, if $P(b)$ is taken as the value from the first piece of evidence and we are updating, $P(b|a_1 \text{ and } a_2) = \epsilon_2$. If $P(b)$ is unknown and is assumed to be 1/2, $P(b|a_1 \text{ and } a_2)$ is again $O(\epsilon)$.

## Discussion of Theorem 3

The pre-step of combining evidence but carrying the evidence as part of the belief function results in a new function with opposing views contained within it. It can still be difficult to determine the net effect on the outcome $b$, especially if one piece of evidence is not considered more important than the other. If a combination of the only defined mass functions is then made, results are obtained which are different from simply combining the evidence directly as in J. Pearl. (Recall that these results would give $Bel\{b\} \cong \frac{\epsilon_2}{\epsilon_1+\epsilon_2}$ and $Bel\{b\}^c \cong \frac{\epsilon_1}{\epsilon_1+\epsilon_2}$). They are instead consistent (of same order) with the probabilistic interpretations given in Pearl[19] and discussed earlier in this paper.

The method actually involves using one frame of reference to translate the uncertain rules to beliefs under a hypothesis of maximizing imprecision and avoiding arbitrariness. This step puts the two rules under one frame of reference and yet maintains their separate identities within this frame. Only two non-zero beliefs result in this frame and they are opposing as they were to begin with. One subtle change has taken place however, which has a small and yet crucial algebraic effect on the **normalizing constant** when the coarser frame is used to re-combine (really combine) the evidence.

In the example of Section 2, there is an overlap between pieces of evidence. The set $\{fp, ofb\}$ appears instead of just $\{ofb\}$. (This would be like using $\{(a_1,b),(a_2,b)\}$ instead of just $\{(a_1,b_)\}$ on the left header side of the table in Theorem 3.) If the set $\{ofb\}$ is used initially with mass function value $1-\epsilon_2$ in Theorem 1 and then the method of Theorem 3 is applied, the results will be as they already are in Theorem 1. Another possibility is to use the belief in $\{fp\}$ and mass on $\{fp, ofb\}$ (not including the mass on $\{ofb\}$ separately). These are really the opposing sets. In this case this case the belief in $\{fp\}$ becomes $O(\epsilon)$ and in $\{fp\}^c$ is of order $1 - O(\epsilon)$.

## §4.3 Conclusions

Methods of obtaining composite belief functions which provide reasonable results consistent with those found using conditional probability theory have been presented for a problem in non-monotonic reasoning. In [19], it is stated that in D-S theory there is the 'unacceptable phenomenon that rule $r_3$, stating that penguins are a subclass of birds, plays no role in the analysis'. This section has argued that unless this information does play a role, the assumptions of the Dempster-Shafer model are not being met and the information is being improperly combined. The problem is handled by choosing a refinement of the frame of discernment which includes the smaller class explicitly. Then a belief function, which is consistent with the coarser frame, is assigned to the finer one. Finally the rule about the subset is incorporated into the refined frame. The combination of the two resulting functions using the usual Dempster rule produces one new function with belief intervals which are intuitively reasonable and more equivalent to the laws of conditional probability. This provides a way to represent non-monotonicity of this type and achieve acceptable results. Section 3 has shown a way to interpret and combine conflicting rules in a belief function framework. The results found are more consistent with those of probability theory. Problems often caused by the normalizing constant do not arise when the methods of Section 3 are employed.

Having said the above, one must still not forget that if the aim is to model the thinking processes of humans, then the belief that penguins fly could well be modified by the knowledge about birds. It is likely that this modification would be small and more consistent with the interpretation given at the end of the discussion to Theorem 3, where $Bel\{fp\}$ becomes $O(\epsilon)$ and $Bel\{fp\}^c$ is $1 - O(\epsilon)$, than with accepting a full conditioning of the belief that birds fly down to the space of penguins.

The methods described in this paper illustrate several comments appearing in [22] by Shafer. To quote, *"the frame of discernment most appropriate to a particular instance of probable reasoning cannot, then, be determined by a priori considerations; it depends on what evidence is available. Notice too, that we will tend to enlarge our frame as more evidence becomes available."* He continues to note that *"the construction of a frame of discernment is a creative act and we choose among frames of discernment by asking not which is* **truer** *but which is more* **useful**. *Furthermore, the translation of our vague knowledge into degrees of support within our frame of discernment can be a challenge to the judgement of our astutest minds"*. These old, but possibly somewhat forgotten comments, I feel are extremely important. Although Dempster's combination rule is fixed, varying the other factors (frame and degrees of support) can greatly alter the results. <u>Providing hard and fast rules for the latter can be dangerous and at best based on assumptions of no absolute validity. This paper has attempted to illustrate these quotes by studying some classical 'bête's noires' of belief function theory and showing how more 'useful' and/or 'intuitive' results can be obtained. However the efficacy of even these suggestions depends largely on the specifics of the application at hand.</u>

**Acknowledgement:** The author wishes to acknowledge discussions with Judea Pearl and P. Smets.